\documentclass[a4paper]{article}

\usepackage{INTERSPEECH2021}
\usepackage[toc,page]{appendix}
\usepackage{hyperref}
\usepackage{amsmath}
\title{Act-Aware Slot-Value Predicting in Multi-Domain Dialogue State Tracking}
\name{Ruolin Su, Ting-Wei Wu, Biing-Hwang Juang}
\address{
  Georgia Institute of Technology, USA
}
\email{\{ruolinsu,waynewu\}@gatech.edu, juang@ece.gatech.edu}

\begin{document}
\maketitle
\begin{abstract}
  As an essential component in task-oriented dialogue systems, dialogue state tracking (DST) aims to track human-machine interactions and generate state representations for managing the dialogue.
  Representations of dialogue states are dependent on the domain ontology and the user's goals. In several task-oriented dialogues with a limited scope of objectives, dialogue states can be represented as a set of slot-value pairs. 
  As the capabilities of dialogue systems expand to support increasing naturalness in communication, incorporating dialogue act processing into dialogue model design becomes essential.
  The lack of such consideration limits the scalability of dialogue state tracking models for dialogues having specific objectives and ontology.
  To address this issue, we formulate and incorporate dialogue acts, and leverage recent advances in machine reading comprehension to predict both categorical and non-categorical types of slots for multi-domain dialogue state tracking. 
  Experimental results show that our models can improve the overall accuracy of dialogue state tracking on the MultiWOZ 2.1 dataset, and demonstrate that incorporating dialogue acts can guide dialogue state design for future task-oriented dialogue systems.
\end{abstract}

\section{Introduction}

    With the rising demand of automatic human-machine interactions for accomplishing service tasks via a natural language dialogue, task-oriented dialogue systems have been developed and widely applied nowadays. 
    Typically, a task-oriented dialogue system consists of four components: automatic speech recognition (ASR), natural language understanding (NLU), dialogue management (DM), and natural language generation (NLG). Dialogue state tracking (DST) is the core function of the DM module which tracks \vphantom{the communication in }human-machine interactions and generates state representations for managing the conversational flow in a dialogue. 
    Specifically, the DST system takes the results of a speech recognizer and a natural language understander, combined with dialogue context as input to predict distributions over a set of pre-defined variables~\cite{Williams2007}\vphantom{[Williams et al., 2007]}. 
    In a task-oriented dialogue, dialogue states are the probability distributions of user's goals in belief space until the current user utterance.
    
    Representations of dialogue states are dependent on the domain ontology and the involved user's goals\vphantom{speech acts}. 
    Table~\ref{tab:exp-dialog} shows an example of state representations within a task-oriented service dialogue between a user and a system for service across train and hotel domains.  
    Among services via a natural language dialogue, some have well-defined objectives to achieve, e.g. reserving hotels or booking train tickets, in which typical dialogue state representations are a set of \textit{(slot, value)} pairs, e.g., \textit{(destination, cambridge)} and \textit{(day, wednesday)} in a train ticket booking service. 
    The objective of DST is therefore to accurately estimate the user’s goals in previous dialogue and to represent them as such slot-value pairs.
    Such slot-value representations are widely used in quite a few task-oriented dialogues\vphantom{for service automation}, such as ATIS~\cite{Hemphill}, DSTC2~\cite{Henderson}, MultiWOZ 2.0/2.1~\cite{Budzianowski,Eric2019}, etc.
    \begin{table}[t]
      \caption{An example of cross-domain dialogue with dialogue state representations and system dialogue acts in MultiWOZ 2.1}
      \label{tab:exp-dialog}
      \centering
      \begin{tabular}{p{.95\linewidth}}
        \toprule
        USER: Hi, I am looking for a train that is going to cambridge and arriving there by 20:45, is there anything like that?\\
        \textit{Dialogue States:}\\
        \textit{TRAIN:	destination=cambridge, arriveby=20:45}\\
        \midrule
        SYSTEM: Where will you be departing from?\\
        USER: I am departing from Birmingham New Street.\\
        \textit{Dialogue States:}\\
        \textit{TRAIN:	destination=cambridge, arriveby=20:45,}\\
        \textit{departure=birmingham new street}\\
        \textit{Dialogue Acts: Inform, Request} \\
        \midrule
        SYSTEM: Can you confirm your desired travel day?\\
        USER: I would like to leave on Wednesday.\\
        \textit{Dialogue States:}\\
        \textit{TRAIN:	destination=cambridge, arriveby=20:45, }\\
        \textit{departure=birmingham new street, day=wednesday}\\
        \textit{Dialogue Acts: Request}\\
        \midrule
        SYSTEM: I have booked your train tickets, and your reference number is a9nhso9y.\\
        USER: Thanks so much. I would also need a place to stay. I am looking for something with 4 stars and has free WiFi.\\
        \textit{Dialogue States:}\\
        \textit{TRAIN:	destination=cambridge, arriveby=20:45, }\\
        \textit{departure=birmingham new street, day=wednesday}\\
        \textit{HOTEL:	stars=4, intenet=yes, type=hotel}\\
        \textit{Dialogue Acts: OfferBooked}\\
        \bottomrule
      \end{tabular}
    \end{table}
    %
    
    In a human dialogue, speech acts are illocutionary actions contained in utterances that change dialogue states~\cite{searle1985speech}. Similarly, those representing the illocutions of utterances in a human-machine dialogue are known as dialogue acts.
    Generally speaking, dialogue acts are defined in domain ontology and serve the functions of conducting particular tasks, having potentials to guide user utterances and enhance the performance of DST as auxiliary inputs. 
    As an example in Table~\ref{tab:exp-dialog}, the \textit{“Request”} act by the system can result in a dialogue state transformation in the \textit{“train”} domain.
    
    Early works on DST regard speech acts as noisy observations of dialogue acts to update dialogue states.
    Assuming fixed domain ontology, a line of generative methods~\cite{DanBohus,Young2013} are proposed to represent dialogue states at each turn by modeling the joint probabilities in a belief space, which costs enormous manual efforts and limits the scalability to multi-domain dialogues. 
    More recent works represent dialogue states as a set of slot-value pairs~\cite{Mrksic,Rastogi}\vphantom{[Neural belief tracker, Schema]}, where discriminative models have proved their capabilities in tracking dialogue states, by modeling DST as multi-task classification~\cite{Wua,Zhong2018,AbhinavRastogiDilekHakkani-Tur2017,Shi2017} or question-answering~\cite{Gao, Gaoa, Zhang, ZhouAmazonAlexaSearch} problems. 
    Recent discriminative DST models estimate user's goals directly from the dialogue context, ignoring the NLU module and dialogue acts. 
    However, using dialogue acts not only helps reason how dialogue states are predicted, but also improves compatibility of the DST model to existing pipeline dialogue systems and scalability to new domains.
    
    To incorporate dialogue acts in discriminative DST models, we propose an act-aware dialogue state tracker (\textbf{ADST}) to predict slot-value pairs for tracking dialogue states with reasoning and high accuracy. 
    We utilize system dialogue acts because they are easier to acquire and are closely relevant to dialogue system design.
    Furthermore, we exploit advances in reading comprehension (RC) to extend our DST models for task-oriented dialogues with more free-form domain ontology.
    Inspired by the RC-based work on DST~\cite{Zhang,ZhouAmazonAlexaSearch}, we formulate the DST problem as predicting values by querying with two types of slots: categorical slots and non-categorical slots. 
    For categorical slots, we implement multiple-choice RC~\cite{Lai2017} to choose from a pre-defined set of limited values, e.g. the \textit{“hotel parking”} slot with a provided value set \textit{\{Yes, No, Don’t Care, None\}}. 
    For non-categorical slots, slots are firstly determined to be either \textit{Don’t Care}, \textit{None} or a span extracted from dialogue context, 
    and then the system takes a span-based RC approach to predict values as probabilities of start and end positions in the dialogue context, e.g. \textit{“20:45”} for the \textit{“train arriveby”} slot. 
    With the formulation of DST as reading comprehension tasks, our model can predict slot-values without pre-defined value sets by extracting values directly from the dialogue context.
    We also take advantage of pre-trained ELMo embeddings~\cite{Peters} to learn word syntax and semantics in dialogue context. 
    In short, our contributions are as follows. (1) We leverage dialogue acts to attend on slots which bring about accuracy improvement on DST; (2) we propose models for categorical and non-categorical slots formulating DST as RC tasks with scalability; 
    (3) we show dialogue acts are related to and have impact on DST via ablation study and attention weight visualization.
    Code will be available at: \url{https://github.com/youlandasu/ACT-AWARE-DST}.
    
\section{Methods}
\subsection{Problem Formulation}

    We denote the tokenized user utterance as $u_t^{usr}$ and the tokenized agent utterance as $u_t^{sys}$ at dialogue turn $t$. The context $C_t$ of the given dialogue at turn $t$ is defined as the concatenation of the previous agent and user utterances, i.e. $C_t = \{u_1^{sys},u_1^{usr},\dots,u_t^{sys},u_t^{usr}\}$. $C_t$ is analogous to a passage in reading comprehension where the model predicts answers. The sequence of system dialogue acts until turn $t$ is $A_t$ = $\{a_1, a_2, \dots, a_t\}$, where $a_i = \{a_i^1, \dots, a_i^{l_i}\}$ represents the number of $l_i$ dialogue acts in turn $i$. For task-oriented dialogues in each domain $d \in D$, the domain ontology defines a set of slots $S^d=S^d_c \cup S^d_n$, where $S^d_c$ and $S^d_n$ are the sets of categorical and non-categorical slots without overlapping.
    
    For categorical slots $S^d_c$, we construct a passage $C_t$, a question $q_{d,s}$ = \{$d$,$s$\} and options $O_{d,s}$ = \{$V^s_c$, \textit{none}, \textit{dont\_care}\} as those in multiple-choice reading comprehension. $V^s_c$ is the set of possible values in each slot $s \in S^d$. 
    Specifically, a question consists of a domain name and a slot name, and options are a list of all possible values with two special values: \textit{none} and \textit{dont\_care}. 
    For non-categorical slots $S^d_n$, the passage and the question would be the same but options are $O_{d,s}$ = \{\textit{span}, \textit{none}, \textit{dont\_care}\}, which substitutes with a span instead of a pre-defined set of values. If the option of \textit{“span”} applies, predicting values from the dialogue context is equivalent to querying passage $C_t$ with $q_{d,s}$ to find the best matched span in the passage.

\begin{figure*}[t]
      \centering
      \includegraphics[width=\linewidth]{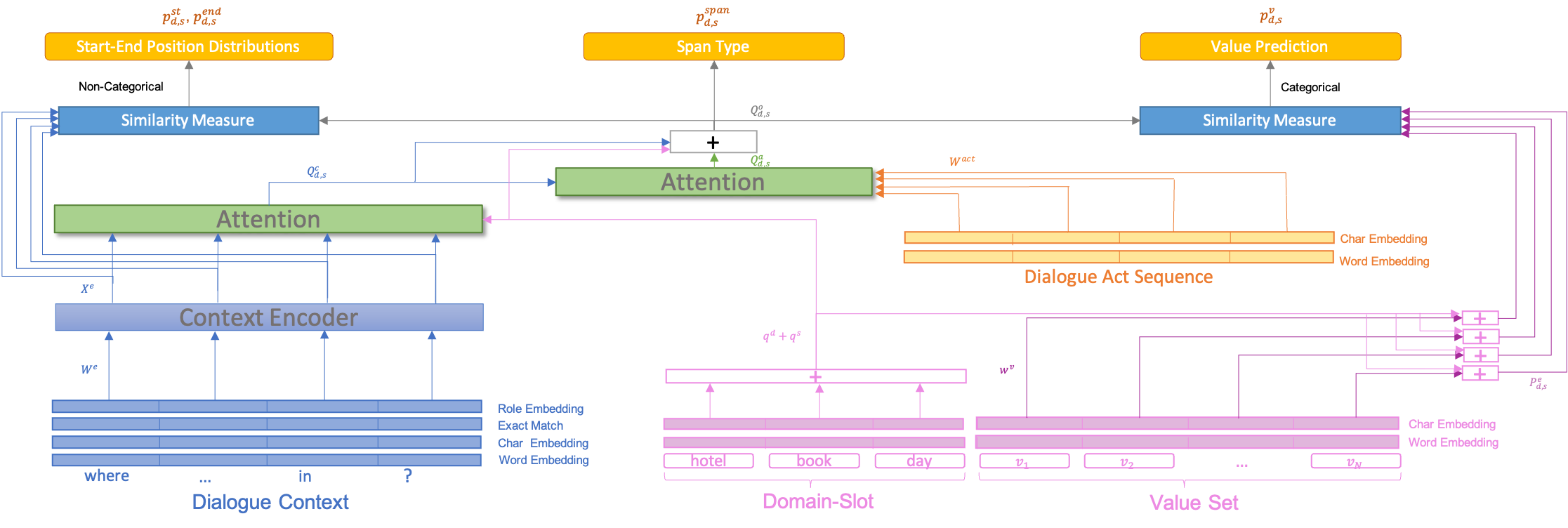}
      \caption{Act-Aware Model Architecture for Dialogue State Tracking. The domain-slot embeddings attend to both encoded dialogue context and previous system dialogue acts. For categorical slots, the similarity scores between possible values and an attended domain-slot is measured to choose from a set of values. For non-categorical slots, a span type is determined and the similarity scores between dialogue context and an attended domain-slot is measured to predict the position of a span.}
      \label{fig:cate-model}
    \end{figure*}

    Figure~\ref{fig:cate-model} shows the overall architecture of our proposed models, mainly consisting of a context encoder, one attention layer attending dialogue context, another attention layer attending system dialogue acts, and two similarity measure modules. 
    
\subsection{Encoding}
    \textbf{Context Encoding}. Denoting the $i$-th word in $C_t$ as $c_i$, we combine word embedding, role embedding and binary exact match features for $c_i$ as the input to the encoder. Specifically, word embeddings $W^e = [W^{ELMo}; W^{Char}]$ are formed by concatenating $W^{ELMo}$ which are ELMo~\cite{Peters} pre-trained word embeddings and $W^{Char}$ which are character-level tokens encoded by CNNs~\cite{Seo}. 
    $W^e\in \mathbb{R}^{|C_t| \times w}$, where $w$ is the sum of word embedding dimension and character embedding dimension while $|C_t|$ is the number of word tokens in the dialogue context. 
    Role embeddings $W^r$ are symbols \textit{“SYS”} or \textit{“USER”} to distinguish the system and the user in a dialogue. 
    Exact match features $W^{exact}$ are binary vectors that reflect where each pre-defined value in the domain ontology is shown in the previous dialogue context. 
    The final input to the encoder is denoted as 
    ${W^e}' = [{W^e}; {W^r}; {W^{exact}}]$. Then we use a bidirectional GRU~\cite{Cho} to encode ${W^e}'$ into $X^e\in \mathbb{R}^{|C_t| \times w}$, i.e. $X^e = BiGRU({W^e}')$, and let the number of bidirectional hidden states be $w$, such that the dimension of encoder’s output is the same as that of $W^e$.
    
    \noindent \textbf{Dialogue Act and Slot-Value Encoding}. 
    The input dialogue act embeddings $W^{act}$ are concatenated word embeddings of system dialogue acts in previous turns.
    We construct similar word embeddings $q^d,q^s \in \mathbb{R}^{w}$ for each domain and slot, respectively. For each dialogue, there are $M$ domain-slot combinations corresponding to $M$ questions. As for categorical slots, we sum $q^d$ and $q^s$ with each value embedding $w^v$ in the value set $V^s_c$ in addition to \textit{“none”}, \textit{“dont\_care”}, and stack them together to construct option embeddings $P_{d,s}^e \in \mathbb{R}^{(N+2) \times {w}}$, i.e. 
    ${P_{d,s}^e}_{i:}= q^d + q^s + w^v_i$, where $N$ is the number of values in the value set $V^s_c$, and $w^v_i\in \mathbb{R}^{w}$ is the $i$-th value in $V^s_c$.  
    
\subsection{Dialogue Context Attention}
     We implement an attention model similar to~\cite{Seo} that computes attention weights between dialogue context and slots. 
     Assuming $R,S$ are two matrices with the same columns $h$, the attention function is defined as:
    \begin{equation}
        Attention_{k}(R,S)_j = softmax_i([R_{i:};S_{j:};R_{i:}\circ S_{j:}]\cdot k)
     \label{eq1}
    \end{equation}
    where $k \in \mathbb{R}^{3h}$ is a trainable vector, $\circ$ is element-wise multiplication, $[;]$ is vector concatenation across column. 
    According to the above definition, the attention weights is computed as $\alpha_{d,s}^{k_1} = Attention_{k_1}(X^e, q^d+q^s)$, and $\alpha_{d,s}^{k_1}\in \mathbb{R}^{|C_t|}$. 
    Then the attended slot vector over dialogue context is $Q_{d,s}^c=(X^e)^T\cdot \alpha_{d,s}^{k_1}$. 
    As a result, $Q_{d,s}^c \in \mathbb{R}^{w}$ is the output of slot embeddings which are dependent on the dialogue context.

\subsection{Dialogue Act Attention}
    In order to fuse information from system dialogue acts, we compute an attended slot vector over dialogue acts
    following Equation~\ref{eq1}. 
    The attention weight of a querying slot attending to acts is given as $\alpha_{d,s}^{k_2} = Attention_{k_2}(W^{act}, Q_{d,s}^c)_{\{d,s\}} \in \mathbb{R}^{|A_t|}$, where $W^{act} \in \mathbb{R}^{ |A_t| \times w}$ is the word embedding of the dialogue acts and $|A_t|$ is the total number of system dialogue acts in previous turns. After that, we obtain a slot vector attended by system acts as $Q_{d,s}^{a}= (W^{act})^T \cdot \alpha_{d,s}^{k_2} \in \mathbb{R}^{w}$.
    
    Then $Q_{d,s}^{a}$ is combined with $Q_{d,s}^c$ and the original slot embedding, i.e. $Q_{d,s}^{o} = Q_{d,s}^c + Q_{d,s}^a + q^d + q^s$, which can be regarded as the final slot embeddings dependent on both the previous dialogue acts and the context. Such that dialogue acts are incorporated in slots.
    

\subsection{Value Classification for Categorical Slots}
    For each categorical slot, a value is to be selected from a pre-defined value set $V^s_c$. 
    Inspired by~\cite{Lai2017}, we compute probability of a value by calculating the bi-linear similarity between possible options $P_{d,s}^e$ and a final slot representations $Q_{d,s}^o$:
    \begin{equation}
        p_{d,s}^v = softmax( P_{d,s}^e \Theta^v  Q_{d,s}^o)
        \label{eq2}
    \end{equation}
   where $\Theta^v$
   is a trainable weight matrix. Denoting $y_{d,s}^v$ as true values for each categorical slots in the dialogue, then the cross entropy loss for the value prediction is calculated as:
    \begin{equation}
        L_v = \sum_t \sum_{d,s} CrossEntropy_t (p_{d,s}^v,y_{d,s}^v)
        \label{eq3}
    \end{equation}

\subsection{Span Prediction for Non-Categorical Slots}
    For non-categorical slots, we first decide the type of span from one of the following options: a span can be extracted from the dialogue context, \textit{“dont\_care”}, or \textit{“none”}. The probability of the span type is calculated by $p_{d,s}^{span} = softmax(FFN_{type}(Q_{d,s}^o))$, where $FFN_{type}$ represents a feed-forward neural network with output dimension of 3. 
    Then we predict the probability distribution of start and end positions in the dialogue context with the following similarity functions: 
    \begin{equation}
    \begin{aligned}
        p_{d,s}^{st} =
        softmax(FFN_{c_1}(X^e) \Theta^s Q_{d,s}^o)
    \end{aligned}
        \label{eq4}
    \end{equation}
    \begin{equation}
        p_{d,s}^{end} = softmax(FFN_{c_2}(X^e) \Theta^e Q_{d,s}^o)
        \label{eq5}
    \end{equation}
    where $FFN_{c_1}$, $FFN_{c_2}$ are one-layer feed-forward networks with output dimensions of $w$, and $\Theta^s, \Theta^e$
    are two trainable weight metrics for predicting the \textit{start} and the \textit{end} position, respectively. Denoting $y_{d,s}^{span}$ as the encoded true label type, the loss function for span type prediction is: 
    \begin{equation}
        L_{type} = \sum_t \sum_{d,s} CrossEntropy_t (p_{d,s}^{span},y_{d,s}^{span})
    \end{equation}
    Then we denote binary vectors of the true start and end positions as $y_{d,s}^{st}$ and $y_{d,s}^{end}$, respectively. The cross entropy loss for predicting span positions is as following:
    \begin{equation}
    \begin{aligned}
        L_s = \sum_t \sum_{d,s} CrossEntropy_t (p_{d,s}^{st},y_{d,s}^{st}) \\
        + \sum_t \sum_{d,s} CrossEntropy_t (p_{d,s}^{end},y_{d,s}^{end})
    \end{aligned}
    \end{equation}
    Finally, the total loss is defined as $L = L_v +  L_{type} +  L_s$.

\section{Experiments} 
\subsection{Dataset}
    We train and evaluate our models on the MultiWOZ 2.1 dataset~\cite{Eric2019}, which is a cross-domain task-oriented dialogue dataset collected from 7 domains containing over 10,000 multi-turn dialogues.
    MultiWOZ 2.1 contains 13 dialogue acts and 30 \textit{(domain, slot)} combinations with hundreds of possible values. We split the dataset into training, development and test set the same as the original setting, and only use 5 most frequent domains in the dataset: \textit{\{restaurant, hotel, train, attraction, taxi\}}.
    
\subsection{Training Details}
    For the input context embeddings, we combine ELMo word embeddings with a length of 512, character embeddings with a length of 100, role embeddings with a length of 128 and one-hot exact matching features indicating occurrences of pre-defined values. 
    For the context encoder, we use a one-layer bi-directional GRU with hidden units of the same length as ELMo embeddings combining character embeddings. 
    ReLu~\cite{nair2010rectified} activation is used for all feed-forward layers.
    The learning rate is 0.001 with the ADAM optimizer~\cite{kingma} and the batch size is 24 in our joint training on 30 categorical and non-categorical slots. 
\subsection{Results}    
    \begin{table}[t]
      \caption{Joint and Slot Goal Accuracy on MultiWOZ 2.1}
      \label{tab:joint-acc}
      \centering
      \resizebox{\columnwidth}{!}{%
      \begin{tabular}{lccc}
        \toprule
        Model & Joint Goal Accuracy & Slot Goal Accuracy\\
        \midrule
        w/o Non-Categorical Slots:\\
        DS-DST picklist~\cite{Zhang}        & 53.30 & -                      \\
        DSTQA w/o span~\cite{ZhouAmazonAlexaSearch} & 51.44   & 97.24\\
        CHAN~\cite{shan2020contextual}  & \textbf{58.55} & \textbf{98.14}\\
        ADST (Ours) all categorical           & 56.70 & 97.71	\\
        \midrule
        w/ Non-Categorical Slots:\\
        STARC~\cite{Gao}         & 49.48   & -                   \\
        DS-DST~\cite{Zhang}        & 51.21 & -                      \\
        DSTQA w/ span~\cite{ZhouAmazonAlexaSearch} & 51.36   & 97.22\\
        ADST (Ours) hybrid & \textbf{56.12} & \textbf{97.62}\\

        \bottomrule
      \end{tabular}
     }
    \end{table}
    Table~\ref{tab:joint-acc} lists the experimental results on MultiWOZ 2.1 test set, where the joint goal accuracy is the average accuracy of predicting all slot-values for a turn correctly, while the slot goal accuracy is the average accuracy of predicting the value of a slot correctly. 
    We compare our models with: (1) DS-DST~\cite{Zhang} which uses BERT-based~\cite{devlin2018bert} RC approaches to handle different slot types jointly, (2) DSTQA~\cite{ZhouAmazonAlexaSearch} which constructs slots with domain ontology for a RC-based model enhanced by a dynamic knowledge graph, (3) CHAN~\cite{shan2020contextual} which fine-tunes BERT in a hierarchical attention network to leverage relevant dialogue context, 
    and 4) STARC~\cite{Gao} which pre-trains on RC dataset then fine-tunes on task dialogues to alleviate data scarcity problems.
    
    We first train a model taking all slots as categorical, comparing it with the categorical-only models: DS-DST picklist, DSTQA w/o span, and CHAN. We achieve 56.70\% joint goal accuracy and 97.71\% slot goal accuracy on categorical-only slot-value predictions, which is close to the state-of-the-art joint and slot goal accuracy on the MultiWOZ 2.1 test set. 
    In contrast to the current state-of-the-art model, our model is lightweight and can be scaled to non-categorical slots.
    In our hybrid model, we take all number- or time-related slots as non-categorical, whereas other slots as categorical, and train all slots jointly. We compare our results with those of STARC, DS-DST and DSTQA w/ span. 
    We obtain outperformed results of 56.12\% and 97.62\% on joint and slot goal accuracy, due to the advantages of exploiting RC approaches and using system dialogue acts as auxiliary inputs.
    
    \begin{table}[h!]
      \caption{Ablation Study on MultiWOZ 2.1 Dev Set}
      \label{tab:ablation}
      \centering
      \resizebox{\columnwidth}{!}{%
      \begin{tabular}{l|c|c}
        \toprule
        Model & Dev Joint & Dev Slot\\
        \midrule
        ADST (Ours) All Categorical   & 56.89 & 97.77  \\
        - w/o Dialogue Acts     & 53.95(-2.94) & 97.53(-0.24) \\
        \midrule
        ADST (Ours) All Non-Categorical   & 45.50 & 96.74 \\
        - w/o Dialogue Acts     & 44.48(-1.02) & 96.65(-0.09) \\
        \bottomrule
      \end{tabular}}
    \end{table}
    \noindent\textbf{Ablation.} We evaluate our models on categorical and non-categorical slots either attending previous system dialogue acts to slots or not.
    Table~\ref{tab:ablation} presents the ablation study results with regard to dialogue acts attention evaluated on the MultiWOZ 2.1 dev set. 
    For our ADST model trained on all categorical slots, removing dialogue act attention layer drops 2.94\% on the joint goal accuracy, and drops 0.24\% on the slot goal accuracy.
    For our ADST model trained on all non-categorical slots, ignoring dialogue acts brings about 1.02\% reduction of joint goal accuracy and 0.09\% reduction of the slot goal accuracy.
    The ablation indicates that our models take advantage from input system dialogue acts for predicting slot-value pairs.
    We also observe that incorporating system dialogue acts into the representations of slots improves the performance of predicting values from the value set by 5\% on joint slot accuracy, but has less impact on span predictions of non-categorical slots.
    
%
    \begin{figure}[h!]
      \centering
      \includegraphics[width=.95\linewidth]{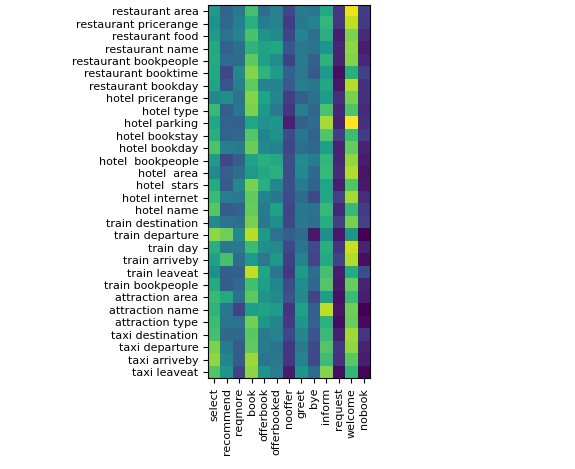}
      \caption{Visualization of attention weights on the dialogue act attention layer of our model for categorical slots.}
      \label{fig:vis}
    \end{figure}
    \noindent\textbf{Visualization of Attention Weights.} To investigate how dialogue acts impact slot-value predictions, we feed a simulated system dialogue act sequence into our trained categorical-only model, and visualize the attention weights between dialogue acts and individual slots in Figure~\ref{fig:vis}. 
    We observe that the \textit{“Request”} act attends the most weights to slots, whereas the \textit{“Welcome”} act attends the least weights. 
    That may be because a \textit{“Request”} act guides a user to give more state-related information in a task-oriented dialogue, but a \textit{“Welcome”} act is not targeted to any dialogue states. 
    Dialogue acts like \textit{“NoOffer”}, \textit{“Reqmore”} yield higher attention weights, probably because they are more targeted to specific dialogue states than more general dialogue acts like \textit{“Select”} or \textit{“Inform”}. 
    Note that the last act usually has comparably higher weights on all slots since it is most relevant to the current dialogue state. 
    Results show that additional attention weights brought by including system dialogue acts suggest correlations between slots and dialogue acts, such that brings about performance improvement on DST. 

\section{Conclusion}
    We propose an act-aware method for multi-domain DST by incorporating system dialogue acts and dialogue context in previous turns to predict slot-value pairs up to the current turn. 
    Our models combine dialogue acts, dialogue context and domain ontology, and leverages reading comprehension approaches to predict slots for both categorical and non-categorical slots. 
    Experimental results show that attentions on both dialogue acts and dialogue context not only improve the joint goal accuracy on MultiWOZ 2.1, but also expand capacities of dialogue systems on reasoning how dialogue states are guided and transformed. 
    The analysis and visualizations indicate that our model is able to use information of system dialogue acts to improve DST on specific slots. 
    We believe that this idea of leveraging dialogue acts in discriminative DST models will improve their scalability for new domains and will contribute to the design of task-oriented dialogue systems for new services.

%

\bibliographystyle{IEEEtran}

\bibliography{mybib}

\begin{thebibliography}{10}
\providecommand{\url}[1]{#1}
\csname url@samestyle\endcsname
\providecommand{\newblock}{\relax}
\providecommand{\bibinfo}[2]{#2}
\providecommand{\BIBentrySTDinterwordspacing}{\spaceskip=0pt\relax}
\providecommand{\BIBentryALTinterwordstretchfactor}{4}
\providecommand{\BIBentryALTinterwordspacing}{\spaceskip=\fontdimen2\font plus
\BIBentryALTinterwordstretchfactor\fontdimen3\font minus
  \fontdimen4\font\relax}
\providecommand{\BIBforeignlanguage}[2]{{%
\expandafter\ifx\csname l@#1\endcsname\relax
\typeout{** WARNING: IEEEtran.bst: No hyphenation pattern has been}%
\typeout{** loaded for the language `#1'. Using the pattern for}%
\typeout{** the default language instead.}%
\else
\language=\csname l@#1\endcsname
\fi
#2}}
\providecommand{\BIBdecl}{\relax}
\BIBdecl

\bibitem{Williams2007}
J.~D. Williams and S.~Young, ``Partially observable markov decision processes
  for spoken dialog systems,'' \emph{Computer Speech \& Language}, vol.~21,
  no.~2, pp. 393--422, 2007.

\bibitem{Hemphill}
C.~T. Hemphill, J.~J. Godfrey, and G.~R. Doddington, ``The atis spoken language
  systems pilot corpus,'' in \emph{Speech and Natural Language: Proceedings of
  a Workshop Held at Hidden Valley, Pennsylvania, June 24-27, 1990}, 1990.

\bibitem{Henderson}
M.~Henderson, B.~Thomson, and J.~D. Williams, ``The second dialog state
  tracking challenge,'' in \emph{Proceedings of the 15th annual meeting of the
  special interest group on discourse and dialogue (SIGDIAL)}, 2014, pp.
  263--272.

\bibitem{Budzianowski}
P.~Budzianowski, T.-H. Wen, B.-H. Tseng, I.~Casanueva, S.~Ultes, O.~Ramadan,
  and M.~Ga{\v{s}}i{\'c}, ``Multiwoz--a large-scale multi-domain wizard-of-oz
  dataset for task-oriented dialogue modelling,'' \emph{arXiv preprint
  arXiv:1810.00278}, 2018.

\bibitem{Eric2019}
M.~Eric, R.~Goel, S.~Paul, A.~Kumar, A.~Sethi, P.~Ku, A.~K. Goyal, S.~Agarwal,
  S.~Gao, and D.~Hakkani-Tur, ``Multiwoz 2.1: A consolidated multi-domain
  dialogue dataset with state corrections and state tracking baselines,''
  \emph{arXiv preprint arXiv:1907.01669}, 2019.

\bibitem{searle1985speech}
J.~R. Searle and D.~Vanderveken, ``Speech acts and illocutionary logic,'' in
  \emph{Logic, thought and action}.\hskip 1em plus 0.5em minus 0.4em\relax
  Springer, 1985, pp. 109--132.

\bibitem{DanBohus}
D.~Bohus and A.~Rudnicky, ``A “k hypotheses+ other” belief updating
  model,'' 2006.

\bibitem{Young2013}
S.~Young, M.~Ga{\v{s}}i{\'c}, B.~Thomson, and J.~D. Williams, ``Pomdp-based
  statistical spoken dialog systems: A review,'' \emph{Proceedings of the
  IEEE}, vol. 101, no.~5, pp. 1160--1179, 2013.

\bibitem{Mrksic}
N.~Mrk{\v{s}}i{\'c}, D.~O. S{\'e}aghdha, T.-H. Wen, B.~Thomson, and S.~Young,
  ``Neural belief tracker: Data-driven dialogue state tracking,'' \emph{arXiv
  preprint arXiv:1606.03777}, 2016.

\bibitem{Rastogi}
A.~Rastogi, X.~Zang, S.~Sunkara, R.~Gupta, and P.~Khaitan, ``Towards scalable
  multi-domain conversational agents: The schema-guided dialogue dataset,'' in
  \emph{Proceedings of the AAAI Conference on Artificial Intelligence},
  vol.~34, no.~05, 2020, pp. 8689--8696.

\bibitem{Wua}
C.-S. Wu, A.~Madotto, E.~Hosseini-Asl, C.~Xiong, R.~Socher, and P.~Fung,
  ``Transferable multi-domain state generator for task-oriented dialogue
  systems,'' \emph{arXiv preprint arXiv:1905.08743}, 2019.

\bibitem{Zhong2018}
V.~Zhong, C.~Xiong, and R.~Socher, ``Global-locally self-attentive dialogue
  state tracker,'' \emph{arXiv preprint arXiv:1805.09655}, 2018.

\bibitem{AbhinavRastogiDilekHakkani-Tur2017}
A.~Rastogi, D.~Hakkani-T{\"u}r, and L.~Heck, ``Scalable multi-domain dialogue
  state tracking,'' in \emph{2017 IEEE Automatic Speech Recognition and
  Understanding Workshop (ASRU)}.\hskip 1em plus 0.5em minus 0.4em\relax IEEE,
  2017, pp. 561--568.

\bibitem{Shi2017}
H.~Shi, T.~Ushio, M.~Endo, K.~Yamagami, and N.~Horii, ``A multichannel
  convolutional neural network for cross-language dialog state tracking,'' in
  \emph{2016 IEEE Spoken Language Technology Workshop (SLT)}.\hskip 1em plus
  0.5em minus 0.4em\relax IEEE, 2016, pp. 559--564.

\bibitem{Gao}
S.~Gao, S.~Agarwal, T.~Chung, D.~Jin, and D.~Hakkani-Tur, ``From machine
  reading comprehension to dialogue state tracking: Bridging the gap,''
  \emph{arXiv preprint arXiv:2004.05827}, 2020.

\bibitem{Gaoa}
S.~Gao, A.~Sethi, S.~Agarwal, T.~Chung, and D.~Hakkani-Tur, ``Dialog state
  tracking: A neural reading comprehension approach,'' \emph{arXiv preprint
  arXiv:1908.01946}, 2019.

\bibitem{Zhang}
J.-G. Zhang, K.~Hashimoto, C.-S. Wu, Y.~Wan, P.~S. Yu, R.~Socher, and C.~Xiong,
  ``Find or classify? dual strategy for slot-value predictions on multi-domain
  dialog state tracking,'' \emph{arXiv preprint arXiv:1910.03544}, 2019.

\bibitem{ZhouAmazonAlexaSearch}
L.~Zhou and K.~Small, ``Multi-domain dialogue state tracking as dynamic
  knowledge graph enhanced question answering,'' \emph{arXiv preprint
  arXiv:1911.06192}, 2019.

\bibitem{Lai2017}
G.~Lai, Q.~Xie, H.~Liu, Y.~Yang, and E.~Hovy, ``Race: Large-scale reading
  comprehension dataset from examinations,'' \emph{arXiv preprint
  arXiv:1704.04683}, 2017.

\bibitem{Peters}
M.~E. Peters, M.~Neumann, M.~Iyyer, M.~Gardner, C.~Clark, K.~Lee, and
  L.~Zettlemoyer, ``Deep contextualized word representations,'' \emph{arXiv
  preprint arXiv:1802.05365}, 2018.

\bibitem{Seo}
M.~Seo, A.~Kembhavi, A.~Farhadi, and H.~Hajishirzi, ``Bidirectional attention
  flow for machine comprehension,'' \emph{arXiv preprint arXiv:1611.01603},
  2016.

\bibitem{Cho}
K.~Cho, B.~Van~Merri{\"e}nboer, C.~Gulcehre, D.~Bahdanau, F.~Bougares,
  H.~Schwenk, and Y.~Bengio, ``Learning phrase representations using rnn
  encoder-decoder for statistical machine translation,'' \emph{arXiv preprint
  arXiv:1406.1078}, 2014.

\bibitem{nair2010rectified}
V.~Nair and G.~E. Hinton, ``Rectified linear units improve restricted boltzmann
  machines,'' in \emph{Icml}, 2010.

\bibitem{kingma}
D.~P. Kingma and J.~Ba, ``Adam: A method for stochastic optimization,''
  \emph{arXiv preprint arXiv:1412.6980}, 2014.

\bibitem{shan2020contextual}
Y.~Shan, Z.~Li, J.~Zhang, F.~Meng, Y.~Feng, C.~Niu, and J.~Zhou, ``A contextual
  hierarchical attention network with adaptive objective for dialogue state
  tracking,'' \emph{arXiv preprint arXiv:2006.01554}, 2020.

\bibitem{devlin2018bert}
J.~Devlin, M.-W. Chang, K.~Lee, and K.~Toutanova, ``Bert: Pre-training of deep
  bidirectional transformers for language understanding,'' \emph{arXiv preprint
  arXiv:1810.04805}, 2018.

\end{thebibliography}

\end{document}